# Nigeria Machinery: A Low-Resource Industrial Dataset with a Domain-Grounded Reasoning Layer

**Gospel Bassey[1], Vincent Fakiyesi[2]**

1. *Independent Researcher*
   *rexkindy@gmail.com*
2. *Engineering Education Transformations Institute, College of Engineering, University of Georgia.*
   *Vincent.Fakiyesi@uga.edu*

Dataset adaptation by Adaption Labs (https://adaptionlabs.ai)

## Abstract

There is relatively little, public, and model-ready data on industrial machinery for African economies. This makes it hard to do quantitative analysis or to train language models on numeric tasks grounded in that setting. We release two things to help with part of this problem.

The first is the Nigeria Machinery Usage and Failures Dataset: 89 machine-level records across 28 indicators, covering Nigeria's manufacturing and oil and gas sectors from 2006 to 2025. Every record names a public source and is decoded by a codebook.

The second is a method for building chain-of-thought (CoT) reasoning examples from these sparse numeric values. The result is 94 prompt, completion, and reasoning-trace rows. In every row, the prompt names the real indicator, subsector, year, and source of the record it comes from. The data adaptation work was carried out by Adaption Labs.

Along the way we describe a problem that is common when language models are used to build datasets. The prompts can match the real numbers while saying nothing about the real domain. We show that fixing this raises the share of domain-grounded prompts from 1 out of 78 in an earlier release to 94 out of 94, and that every retrieval answer now matches its source value (84 out of 84). We release the data, the reasoning layer, and a per-row provenance file under CC-BY-4.0.

We are clear about the limits. With 89 records and 17 indicators that have only one observation, this is a reference and seed dataset, not a large training set. Most reasoning rows are retrieval rather than multi-step computation.

---

# 1. Introduction

Predictive maintenance and fault diagnosis depend on machine-level data: failure counts, downtime, capacity utilization, and maintenance spend, recorded over time. For most of Nigeria's industrial base, this data is not available in any form a model can use. The figures exist, but they are locked inside annual reports as static tables, held in regulator filings, or buried in national statistics that aggregate away everything specific. An analyst cannot easily see how a single refinery performed across a decade, only what a whole sector reported in a given year.

This is a consequential gap. Nigeria is the largest oil producer in Africa, and the sector provides most government revenue and most foreign exchange earnings. When equipment fails, the effects reach past the plant into energy prices and public budgets. Yet there is almost no public, machine-readable record of how that failure unfolds over time, which means the data-driven methods now standard in industrial machine learning cannot be applied to the region at all.

The problem is not only that data is scarce. It is also that the scarcity has a particular shape. The numbers that exist are scattered, inconsistently reported, and never assembled into a single structured resource. Building one requires manual collection from many separate official documents, careful cross-checking, and a consistent encoding so that a model can ingest it. This is slow, unglamorous work, and it is the reason the resource does not yet exist.

This paper takes a first step. We make three contributions:

1. We construct and release the Nigeria Machinery Usage and Failures Dataset, a structured, fully-sourced collection of 89 machine-level industrial records for Nigeria spanning 2006 to 2025, assembled by hand from official government and industry sources (Section 3).
2. We describe a method for turning these sparse tabular values into domain-grounded chain-of-thought reasoning pairs for language-model fine-tuning, carried out with Adaption Labs, and we name and measure a grounding failure mode that this method corrects (Section 4).
3. We release lightweight, reproducible evaluation tooling so that the dataset's reported properties can be verified and any model can be scored against it (Section 5 and Appendix B).

We are deliberate about scope. The dataset is small, and we do not present it as a predictive-maintenance benchmark; it cannot support that at scale. We present it as a reference and seed resource, and as a documented, reusable method that others can apply to similarly data-poor settings.

# 2. Related Work

**Predictive maintenance and industrial fault diagnosis.** Data-driven predictive maintenance is now a large field, with extensive reviews covering condition monitoring, fault diagnosis, and remaining-useful-life estimation [1, 2]. These methods are powerful but data-hungry, and surveys repeatedly note that results validated on controlled-experiment or simulated data often do not carry over to messy real-world deployment [3]. Our work addresses an upstream bottleneck for this field in the African context: the absence of usable data in the first place.

**Industrial fault and maintenance datasets.** Most public benchmarks for machine fault diagnosis come from test rigs or industrial sites in high-income countries. Widely used examples include the Case Western Reserve University (CWRU) bearing dataset and the Paderborn University bearing dataset [4, 5]. They offer scale and good sensor data, but they carry the operating conditions and machine types of those regions, which limits how well they transfer to other settings. Recent work has shown that even within these benchmarks, common evaluation setups can produce over-optimistic results that do not generalize [6]. Synthetic industrial datasets such as AI4I 2020 partly fill the gap, but they simulate rather than record real operating conditions [7]. Our dataset is different in kind: much smaller, but specific to Nigerian sectors and facilities at the indicator level, and drawn from real reported figures.

**Low-resource dataset construction.** Recent work in African and other low-resource NLP creates value by building small, well-documented, region-specific datasets and releasing them openly with strong provenance. The Masakhane participatory-research effort is a prominent example, producing translation benchmarks for over 30 African languages through open community collaboration [8]. The broader practice of documenting datasets carefully, for instance through structured datasheets, is now an established norm [9]. Closest to our approach is recent work that builds chain-of-thought instruction datasets for low-resource settings using an automated, reusable pipeline, such as TIBSTC-CoT for Tibetan [10]; we apply a comparable idea to industrial tabular data rather than language.

**LLM-generated training data and its problems.** Using language models to synthesize instruction or reasoning data is now common [11], but it brings known risks. Chain-of-thought reasoning that is not properly grounded is prone to hallucination [12], and a distinct failure mode, faithfulness hallucination, occurs when a model's output is not supported by its given input or context [13]. Synthetic datasets are also often narrow, tailored to a specific task, which limits how well they transfer [11], and recent frameworks add explicit quality-control stages precisely because LLM-generated records can be implausible or domain-inconsistent [14]. Part of our contribution is to name and measure one of these problems in a real dataset, where prompts matched the right numbers but not the right domain, and to show a fix.

# 3. The Nigeria Machinery Usage and Failures Dataset

## 3.1 Contents

The core dataset has 89 records across 28 indicators, covering 2006 to 2025. Each record is one coded observation with year, period, sector, subsector, indicator, value, unit, and source. All fields are integer-coded and resolved by a codebook.

- **Sectors:** Industrial Manufacturing (38 records) and Oil and Gas (51 records).
- **Subsectors (6):** General Manufacturing; the Port Harcourt, Warri, and Kaduna refineries; Upstream and Gas Flaring; and Crude Oil Production.
- **Indicators (28):** capacity utilization, refinery capacity utilization, downtime, maintenance and turnaround spend, shutdown events, energy costs, production output, gas flared, jobs added and lost, GDP share, and more.

Table 1 summarizes the composition of the dataset by subsector, showing how many core records each subsector contributes and how many reasoning-layer rows are derived from them.

| Subsector | Core Records | Reasoning Rows |
| --- | --- | --- |
| General Manufacturing | 38 | 43 |
| Port Harcourt Refinery | 27 | (in oil and gas total) |
| Warri Refinery | 10 | (in oil and gas total) |
| Kaduna Refinery | 8 | (in oil and gas total) |
| Upstream and Gas Flaring | 2 | (in oil and gas total) |
| Crude Oil Production | 4 | (in oil and gas total) |
| **Manufacturing total** | **38** | **43** |
| **Oil and Gas total** | **51** | **51** |
| **All** | **89** | **94** |

Table 1: Dataset composition by subsector. The reasoning layer adds derived multi-record computation rows, which is why the oil and gas reasoning count slightly exceeds a one-to-one mapping with core records.

## 3.2 Data collection

The records were collected manually. The starting point was the published documents of official Nigerian institutions: the Central Bank of Nigeria (CBN) Statistical Bulletin and statistics database, which publishes manufacturing capacity utilization rates and lending rates; the National Bureau of Statistics (NBS), for GDP and manufacturing-output figures; the Nigerian Upstream Petroleum Regulatory Commission (NUPRC) annual reports, for upstream production, flaring, and incident data; the Manufacturers Association of Nigeria (MAN) economic reviews; and published NNPC financial statements. Where an official primary source did not cover a specific figure, we used reputable secondary sources that themselves cite the underlying data, including industry analyses and academic case studies, and we recorded the secondary source rather than implying primary access.

For each figure, the process was as follows. We located the value in a source document, most often a table inside a PDF annual report or statistical bulletin. We then cross-checked it: where the same indicator appeared in more than one source, we compared the figures and preferred the official or government source when they agreed, and recorded the discrepancy when they did not. Official and government sources were preferred throughout, and the specific source for every record is stored in the source_code field rather than left implicit. This means a reader can trace any single value back to the document it came from.

We did not attempt to fill gaps by estimation or interpolation. If a figure for a given indicator, facility, and year could not be found in a source we trusted, we left it out. This is the main reason the dataset is small and unevenly covered (Section 3.4): it reflects what is actually documented and verifiable, not what could be inferred.

### 3.3 Encoding and the codebook

Once collected, the raw figures were encoded into a consistent integer-coded schema so that the dataset could be ingested directly by a model without free-text parsing. Each record holds coded fields for year, period, sector, subsector, indicator, unit, and source, plus the numeric value. The coding scheme and the accompanying codebook, which maps every integer code to its real-world meaning, were designed jointly by the author and Adaption Labs. The codebook is released with the dataset so that the encoding is fully reversible and self-documenting.

### 3.4 Coverage and its unevenness

Coverage is uneven, and we report it that way on purpose. Two indicators hold most of the time depth: capacity utilization and refinery capacity utilization. Seventeen of the 28 indicators have only one record. The practical result is simple. The dataset supports reference lookup, descriptive and longitudinal analysis on the well-covered indicators, and the seeding of reasoning tasks. It is not enough on its own to train a predictive-maintenance failure classifier at scale. For that, combine it with broader industrial data.

### 3.5 A worked signal

Even at this size, the data shows real structure. General Manufacturing capacity utilization sits in the low-to-mid 50s (percent) through the 2010s, drops to about 40 percent in 2020, then recovers. That dip lines up with the pandemic. The refineries tell a harder story, with single-digit utilization in several years, which matches the well-documented near-dormancy of Nigerian refining before recent rehabilitation work.

## 4. Method: Domain-Grounded Chain-of-Thought from Tabular Values

The reasoning layer was built with Adaption Labs, who carried out the data adaptation.

### 4.1 The grounding problem

An earlier public release paired the records with 78 LLM-generated prompt, completion, and trace rows for fine-tuning. When we looked closely, the numeric answers were real and each one traced back to a record value. But the prompts were generic. A real value of 35.5 was wrapped as a medical-sensor temperature conversion. Another was wrapped as an abstract log base 2 of 512. Out of 78 prompts, only 1 mentioned any industrial term.

So the data was grounded in the right numbers, but not in the right meaning. A prompt about medical sensors teaches a model nothing about Nigerian refinery operations. This is the problem we set out to fix, and we report it openly because it is easy to introduce and easy to miss.

## 4.2 The fix

We rebuild the reasoning layer straight from the decoded records. For each record, the prompt names the real indicator, subsector, year, and source. The style depends on the indicator type:

- **Natural-language retrieval** for level and rate indicators. For example, a reliability engineer asking for a refinery's capacity utilization in a given year, citing the real source.
- **Strict structured** prompts for indicators given as changes, costs, or shares, where a rigid format keeps the numeric task clear.
- **Classification** framing for the binary shutdown-event indicator.

We vary the persona (reliability engineer, maintenance planner, control-systems analyst, operations economist) to keep the model from overfitting one style. The numeric answer is never invented. It is the record value, or a correct arithmetic result over real record values.

## 4.3 Genuine multi-value computations

Where the indicators allow it, we build real multi-step computations instead of lookups. These are the rows that test reasoning rather than recall:

- Residual downtime after a reported predictive-maintenance reduction, using the real reported reduction figure.
- Cost per downtime day, computed two independent ways that cross-check to within rounding.
- Equipment-failure and human-error shares of downtime, using the real reported shares.
- A capacity-utilization change between two real consecutive readings.

## 4.4 Generation pipeline and acceptance gate

The full procedure runs as a fixed sequence of stages, applied to every record:

1. **Collect and decode.** Start from a real core record and resolve its codes (indicator, subsector, year, unit, source) into plain terms using the codebook.
2. **Select style.** Choose the prompt style from the indicator type: natural-language retrieval, strict structured, or classification (Section 4.2).
3. **Generate prompt.** Build a domain-grounded prompt that names the real indicator, subsector, year, and source, with a persona drawn from a fixed pool.
4. **Generate answers and trace.** Set the answer to the real record value, or to a correct arithmetic result over real values, and write a reasoning trace that shows the steps over those real values.
5. **Check against an acceptance gate.** Apply automatic acceptance tests before a row is admitted (described below).
6. **Audit.** Record the row's source record or records, operation, and answer in a provenance file.

The acceptance gate in step 5 is explicit, and a row is admitted only if it passes every test: the prompt must contain at least one domain term that ties it to the industrial setting; for retrieval rows, the stated answer must match the source record value to the dataset's precision; for computation rows, the answer must equal the recomputed arithmetic result over the cited real values; and the row must trace to at least one real core record. Rows that fail any test are not silently dropped but corrected and re-checked, so the released set is the set that passes the gate. The metrics in Section 5.1 report the pass rates of this gate, which is why they read as high: they are not a sample, but the acceptance criterion applied to the whole layer.

### 4.5 Output and audit

The method produces 94 rows: 59 natural-language retrieval, 25 strict retrieval, 6 genuine computations, and 4 shutdown classifications. Each row comes with a provenance entry that links it to its source record or records. We release this audit file so anyone can check the grounding and the answer of every row.

## 5. Evaluation

We build the evaluation around what 89 records can honestly support. That means the integrity of the reasoning layer, not a competition result.

### 5.1 Grounding and answer-fidelity metrics

All the numbers below can be reproduced from the released audit file.

| Property | Earlier Release | After the fix |
|---|---|---|
| Domain-grounded prompts | 1/78 | 94/94 |
| Retrieval completion matches source value | 7 mismatches present | 84/84 |
| Prompts naming a real year | Not applicable | 93/94* |
| Genuine multi-step computations | 1 | 6 |
| Rows traceable to a real record | 78/78 | 94/94 |

*The one exception is a period-over-period change record, where an absolute year-value framing does not apply.

Table 2: Grounding and answer-fidelity metrics, before and after re-grounding. Each "after" figure is the pass rate of the acceptance gate described in Section 4.4, applied to the full reasoning layer.

### 5.2 Suggested fine-tuning protocol

For users who want to fine-tune, we suggest treating the retrieval rows (84) and the computation rows (6) as separate strata. They test different abilities. Retrieval tests provenance-grounded recall. Computation tests multi-step numeric reasoning. If you mix them, you will overstate reasoning performance.

A small instruction-tuning run on the reasoning layer should teach the strict numeric-output format and the habit of citing the source. It will not teach general industrial reasoning from 6 computation examples. We say this so the dataset is used for what it can actually do.

# 6. Limitations

We list these on purpose, and we treat stating them as part of the contribution.

- **Scale.** The dataset has 89 records, and 17 indicators have only one observation. This is a seed and reference set, not a training set.
- **Retrieval versus reasoning.** Most reasoning rows are retrieval. Only 6 are real multi-step computations. Measure them separately.
- **Source heterogeneity.** Some values come from industry or advocacy reports, not audited statistics. For example, a reported 83 percent predictive-maintenance downtime reduction, used in one computation, is a vendor-study figure and should be read as a claim, not a measurement. The per-record source code is there so users can apply this judgment.
- **Templated phrasing.** The values and provenance in the prompts are real, but the wording is template-generated and persona-varied. It is not a record of how practitioners actually phrase queries.
- **Single annotator.** One author compiled the data, with adaptation support from Adaption Labs. The dataset has not had independent inter-annotator checking.

# 7. Ethics and Responsible Use

The dataset has no personal data. All records are sector-level or facility-level economic and operational indicators from public sources. Because some figures come from non-audited reports, we caution against using the dataset for high-stakes operational or policy decisions without corroboration, and against presenting model outputs trained on it as if they were measurements. The provenance file is there to make responsible weighting possible.

# 8. Availability

The dataset, the reasoning layer, the codebook, and the per-row provenance file are released under CC-BY-4.0 on Hugging Face and Kaggle. The reasoning layer can be reproduced from the released records and the audit file.

- Hugging Face: https://huggingface.co/datasets/gospelgit/Nigeria_Machinery_Dataset (profile: https://huggingface.co/gospelgit)
- Kaggle: https://www.kaggle.com/datasets/gospelnbassey/n-m-machinery

# 9. Conclusion

We have described a small, real, fully-sourced dataset of Nigerian machine-level industrial indicators, and a method for building domain-grounded chain-of-thought data from sparse numeric values. The main point carries beyond this dataset: matching the numbers is not the same as matching the meaning, and the gap between them can be measured and fixed. We hope the dataset seeds more work on African industrial data, and that the method, along with the habit of auditing it, is useful to others building reasoning data from small tabular sources.

# Acknowledgements

Data adaptation was carried out by Adaption Labs (https://adaptionlabs.ai).

# References

[1] Y. Lei, N. Li, L. Guo, N. Li, T. Yan, and J. Lin. Machinery health prognostics: A systematic review from data acquisition to RUL prediction. *Mechanical Systems and Signal Processing*, 104:799-834, 2018. doi:10.1016/j.ymssp.2017.11.016.

[2] Y. Lei, B. Yang, X. Jiang, F. Jia, N. Li, and A. K. Nandi. Applications of machine learning to machine fault diagnosis: A review and roadmap. *Mechanical Systems and Signal Processing*, 138:106587, 2020. doi:10.1016/j.ymssp.2019.106587.

[3] T. P. Carvalho, F. A. A. M. N. Soares, R. Vita, R. da P. Francisco, J. P. Basto, and S. G. S. Alcalá. A systematic literature review of machine learning methods applied to predictive maintenance. *Computers & Industrial Engineering*, 137:106024, 2019. doi:10.1016/j.cie.2019.106024.

[4] W. A. Smith and R. B. Randall. Rolling element bearing diagnostics using the Case Western Reserve University data: A benchmark study. *Mechanical Systems and Signal Processing*, 64-65:100-131, 2015. doi:10.1016/j.ymssp.2015.04.021.

[5] C. Lessmeier et al. Condition monitoring of bearing damage in electromechanical drive systems by using motor current signals of electric motors: A benchmark data set for data-driven classification. *PHM Society European Conference*, 3(1), 2016. doi:10.36001/phme.2016.v3i1.1577.

[6] R. K. Rosa et al. Benchmarking deep learning models for bearing fault diagnosis using the CWRU dataset: A multi-label approach. *arXiv preprint*, arXiv:2407.14625, 2024.

[7] S. Matzka. Explainable artificial intelligence for predictive maintenance applications. *Third International Conference on Artificial Intelligence for Industries (AI4I)*, pages 69-74, 2020. doi:10.1109/AI4I49448.2020.00023. (AI4I 2020 Predictive Maintenance dataset.)

[8] W. Nekoto et al. (Masakhane). Participatory research for low-resourced machine translation: A case study in African languages. *Findings of the Association for Computational Linguistics: EMNLP 2020*, pages 2144-2160, 2020. doi:10.18653/v1/2020.findings-emnlp.195.

[9] T. Gebru et al. Datasheets for datasets. *Communications of the ACM*, 64(12):86-92, 2021. doi:10.1145/3458723.

[10] F. Gao, C. Huang, N. Tashi, et al. TIBSTC-CoT: A multi-domain instruction dataset for chain-of-thought reasoning in language models. *arXiv preprint*, arXiv:2508.01977, 2025.

[11] J. Wei et al. Chain-of-thought prompting elicits reasoning in large language models. *Advances in Neural Information Processing Systems*, 35:24824-24837, 2022.

[12] A. Plaat et al. Multi-step reasoning with large language models, a survey. *ACM Computing Surveys*, 2025. doi:10.1145/3774896.

[13] L. Huang et al. A survey on hallucination in large language models: Principles, taxonomy, challenges, and open questions. *ACM Transactions on Information Systems*, 43(2), Article 42, 2025. doi:10.1145/3703155.

[14] (Team-then-Trim) Team, then trim: An assembly-line LLM framework for high-quality tabular data generation. *arXiv preprint*, arXiv:2602.04785, 2026.

---

# Appendix A. Indicator List

The 28 indicators, with units, are resolved in the released codebook. Examples include capacity utilization rate (percent), refinery capacity utilization rate (percent), average unplanned downtime (days per year), refinery maintenance and debt cost (billion NGN), rehabilitation and turnaround spend (million USD), refinery shutdown event (binary), natural gas flared (BCF), average daily crude oil production (mbpd), and manufacturing sector share of GDP (percent).

# Appendix B. Reproducibility

Every reasoning-layer row can be regenerated from the core records using the documented per-indicator templates. Every row's source record or records, operation, and answer are listed in the audit file. The grounding and answer-fidelity numbers in Section 5.1 are computed directly from that file.

We release two lightweight scripts (pandas only, no GPU required) alongside the data:

- evaluate_dataset.py reproduces the Section 5 metrics directly from the three released files. Running it prints the domain-grounding, year-naming, retrieval-fidelity, computation-count, and traceability figures reported in this paper.
- eval_harness.py is a model-agnostic scorer. It exports the prompts for any model to answer, then scores the returned predictions for numeric exact-match and format compliance, reporting retrieval, computation, and classification strata separately so that retrieval recall does not mask reasoning performance.

Together these let a reader verify the reported metrics and evaluate any model on the dataset without specialized hardware.